\documentclass[review]{elsarticle}

\usepackage{lineno,hyperref}
\usepackage{graphicx}
\modulolinenumbers[5]
\usepackage{multirow}
\usepackage[table,xcdraw]{xcolor}
\journal{Journal}

\usepackage{hyperref}
\usepackage[colorinlistoftodos]{todonotes}
\hypersetup{
    colorlinks=true,
    linkcolor=blue,
    filecolor=magenta,      
    urlcolor=cyan,
}

\begin{document}

\begin{frontmatter}

\title{BeCAPTCHA-Mouse: Synthetic Mouse Trajectories \\ and Improved Bot Detection}

\author{Alejandro Acien\corref{mycorrespondingauthor}}\ead{alejandro.acien@uam.es}
\author{Aythami Morales\corref{mycorrespondingauthor}}\ead{aythami.morales@uam.es}
\author{Julian Fierrez}\ead{julian.fierrez@uam.es}
\author{Ruben Vera-Rodriguez}\ead{ruben.vera@uam.es}
\address{Biometrics and Data Pattern Analytics Lab, Universidad Autonoma de Madrid, Spain}

\cortext[mycorrespondingauthor]{Corresponding author}

\begin{abstract}
We first study the suitability of behavioral biometrics to distinguish between computers and humans, commonly named as bot detection. We then present BeCAPTCHA-Mouse, a bot detector \textcolor{black}{based on: \textit{i)} a neuromotor model of mouse dynamics to obtain a novel feature set for the classification of human and bot samples; and \textit{ii)}  a learning framework involving real and synthetically generated mouse trajectories.} We propose two new mouse trajectory synthesis methods for generating realistic data: \textit{a)} a function-based method based on heuristic functions, and \textit{b)} a data-driven method based on Generative Adversarial Networks (GANs) in which a Generator synthesizes human-like trajectories from a Gaussian noise input. Experiments are conducted on a new testbed also introduced here and available in GitHub: BeCAPTCHA-Mouse Benchmark; useful for research in bot detection and other mouse-based HCI applications. Our benchmark data consists of $15$,$000$ mouse trajectories including real data from $58$ users and bot data with various levels of realism. Our experiments show that BeCAPTCHA-Mouse is able to detect bot trajectories of high realism with $93\%$ of accuracy in average using only one mouse trajectory. When our approach is fused with state-of-the-art mouse dynamic features, the bot detection accuracy increases relatively by more than $36\%$, proving that mouse-based bot detection is a fast, easy, and reliable tool to complement traditional CAPTCHA systems.
\end{abstract}


\end{frontmatter}

\nolinenumbers 
\section{Introduction}
\label{intro}

\textcolor{black}{During the last decades, the security applications have had a key role in the development of machine learning technologies. Thus, research areas such as fingerprint identification, face recognition, iris recognition, or person re-identification have attracted the interest of the research community promoting continuous advances in their fields. These advances resulted in more accurate physical security systems and advances in state-or-the-art. However, security threats are moving from the physical domain to the digital domain. The Cybercrime is increasing in both percentage of citizens affected and cost in the global economy\footnote{\url{https://www.cbronline.com/news/cybercrime-cost-fbi}}. The criminals become more and more sophisticated and the cross-border nature of a large percentage of these crimes difficult the fight. The challenges and potential benefits of technologies developed to serve in this fight are large and the Pattern Recognition community can play an important role in this scenario. Among these challenges, the present work is focused on the detection of bots and how pattern recognition techniques and machine learning frameworks can be used to develop new approaches.}  

How to distinguish between human users and artificial intelligence during computer interactions is not a trivial task. This challenge was firstly discussed by Alan Turing in 1950. He investigated whether machines could show an intelligent behavior, and also how humans could be aware of these artificial behaviors. For this, he developed the famous Turing Test, commonly named as \textit{The Imitation Game}, in which a human evaluator would judge natural language conversations between a human and a computer designed to generate human-like responses. The Turing Test was both influential and widely criticized and became an important concept in the artificial intelligence field \cite{Saygin}. However, at the epoch of Alan Turing research, the problem of machines acting like humans were commonly associated to science-fiction topics.

Nowadays, boosted by the last advances of machine learning technologies and worldwide connections, that ‘science-fiction topic’ becomes a real hazard. As an example, bots are expected to be responsible for more than $40\%$ of the web traffic with more than $43\%$ of all login attempts to come from malicious botnets in the next years\footnote{\hyperlink{https://resources.distilnetworks.com/white-paper-reports/bad-bot-report-2019}{https://resources.distilnetworks.com/white-paper-reports/bad-bot-report-2019}}. Malicious bots cause billionaire losses through web scraping, account takeover, account creation, credit card fraud, denial of service attacks, denial of inventory, and many others. Moreover, bots are used to influence and divide society (e.g. usage of bots to interfere during Brexit voting day \cite{Gorodnichenko}, or to spread anxiety and sadness during the COVID-19 outbreak\footnote{\hyperlink{www.washingtonpost.com/science/2020/03/17/analysis-millions-coronavirus-tweets-shows-whole-world-is-sad/}{https://www.washingtonpost.com/science/2020/03/17/analysis-millions-coronavirus-tweets-shows-whole-world-is-sad/}}$^{,}$\footnote{\hyperlink{https://www.sciencealert.com/bots-are-causing-anxiety-by-spreading-coronavirus-misinformation}{https://www.sciencealert.com/bots-are-causing-anxiety-by-spreading-coronavirus-misinformation}} through Twitter). Bots are becoming more and more sophisticated, being able to mimic human online behaviors. On the other hand, algorithms to distinguish between humans and bots are also getting very complex. We can distinguish two types of bot detection methods in response to those sophisticated bots:

\begin{itemize}
\setlength\itemsep{0em}
\item Active Detection. Traditionally named as CAPTCHA (Completely Automated Public Turing test to tell Computers and Humans Apart), these algorithms determinate whether or not the user is human by performing online tasks that are difﬁcult for software bots to solve while being easy for legitimate human users to complete. Some of the most popular CAPTCHA systems are based on: characters recognition from distorted images (text-based), class-objects identification in a set of images (image-based), and speech translation from distorted audios (audio-based).
\item Passive Detection. These detectors are transparent and analyze the users behavior while they interact with the device. The last version of Google reCAPTCHA v3 replaces traditional cognitive tasks by a transparent algorithm capable of detecting bots and humans from their web behavior\footnote{\hyperlink{https://www.google.com/recaptcha/intro/v3.html}{https://www.google.com/recaptcha/intro/v3.html}}. Other researchers \cite{Xie}, describe browsing behavior of web users for detection of DDoS Attacks (Distributed Denial of Service).
\end{itemize}

Although these algorithms are broadly used, they present limitations. First of all, ensuring a very accurate bot detection makes the tasks difficult to perform even for humans.  Second, most of the CAPTCHA systems can be easily solved by the most modern machine learning techniques. For example, the text-based CAPTCHA was defeated by Bursztein \textit{et al.} \cite{Bursztein} with $98\%$ accuracy using a ML-based system to segment and recognize the text. In \cite{Kevin}, the authors designed an AI-based system called unCAPTCHA to break Google’s most challenging audio reCAPTCHAs. \textcolor{black}{The last version of the Google CAPTCHA, named reCAPTCHAv3, was systematically fooled in \cite{Ismail} by synthesizing mouse trajectories using reinforcement learning techniques.} Third, these algorithms process sensitive information and there are important concerns about how they comply with new regulations such as the European GDPR\footnote{\hyperlink{https://complianz.io/google-recaptcha-and-the-gdpr-a-possible-conflict/}{https://complianz.io/google-recaptcha-and-the-gdpr-a-possible-conflict/}}. Fourth, the CAPTCHA systems become a great barrier to people with visual or other impairments. Finally, the Turing Test was designed as a task in which machines had to prove they were human, meanwhile in current CAPTCHA systems humans have to prove they are not machines (e.g. \textit{I’m not a robot} from Google’s). This means that the responsibility to prove the user’s ‘humanity’ falls over human users instead of bots. At this point, there is still a large room for improvement towards reliable bot detection able to stop malicious software not bothering human users during natural web browsing.

On the other hand, \textcolor{black}{Machine Learning and Pattern Recognition communities have made great advances during the last decades. These advances have boosted several research fields including Computer Vision, Audio Processing, and Natural Language Processing. Nonetheless, the application of these advances to the bot detection field has been rather low. While previous works \cite{Bursztein,Kevin} focus their efforts in beating the existing CAPTCHA systems and exposing their vulnerabilities with the latest advances in machine learning techniques, we use them to develop better bot detectors and harden the existing ones. Among the different technologies proposed during the last years, this work proposes to improve bot detection using the progress made in two specific areas: \textit{i)} Behavioral Biometrics \cite{picard2020behavior, Javier_2020}, and \textit{ii)} Generative Adversarial Networks (GANs).} 

Biometric recognition refers to the automated recognition of individuals based on their physiological (e.g. fingerprint, face) and behavioral (e.g. keystroke, gait) characteristics \cite{Anil}. Traditionally focused on person recognition, the individual patterns obtained from biometric signals characterize the human being. Behavioral biometrics refers to those traits revealing distinctive user behaviors and mannerisms when they interact with devices (e.g. smartphones, tablets) \cite{2019_MULEA_Acien_MultiLock}. Behavioral biometrics characteristics can be easily acquired with almost total transparency, being less invasive than other methodologies. \textcolor{black}{In the biometrics research literature, most works so far for securing services and systems against attacks \cite{Galbally2007_Vulnerabilities} have been focused either in template protection based on cryptographic constructions \cite{2017_Access_HEmultiDTW_Marta} or liveness detection against presentation attacks \cite{hadid15SPMspoofing}. Utilizing behavioral-based biometrics for improving the security against bots and other kind of attacks has been only studied very timidly \cite{2019_BookPAD2_IntroSignPAD_Tolosana}. Some examples in this regard using behavioral features to train cognitive models to parameterize the user behavior and detect patterns useful to improve the security of digital services can be found in \cite{Lidia, Ogiela}.}

\textcolor{black}{On the other hand, Generative Adversarial Networks (GANs) appeared in 2014 as a data-driven method for generating synthetic samples from real ones \cite{2020_JSTSP_GANprintR_Neves}. Since then, GANs have shown impressive improvements over previous generative methods, such as variational auto-encoders or restricted Boltzmann machines. The GAN architecture consists of two networks trained together in an adversary manner: the Generator and the Discriminator. While the Generator generates synthetic data by learning the statistical distribution of real data, the Discriminator is a classifier that discriminates whether an input is real or synthetically generated. GANs have been widely used in many applications, especially in the generation of synthetic images. The application of GANs to the generation of synthetic time signals is much scarcer. One of the first approaches on the use of GANs for time series generation was done by Morgren \cite{C-RNN-GAN_2016}, utilizing both recurrent neural networks (RNN) and GANs to synthesize music data. Other works presenting novel architectures for the generation of time signals using GANs are \cite{esteban2017realvalued, NEURIPS2019_TimeGans}.}

\textcolor{black}{Behavioral biometrics and GANs have been applied successfully in bot detection for mobile devices scenarios \cite{2021_EAAI_BeCAPTCHA_Acien}. The method proposed in \cite{2021_EAAI_BeCAPTCHA_Acien} combines information from the accelerometer and touchscreen sensors. However, in that work the software-based sampling rate of mobile devices and the simplicity of touch over touchscreens limited the results. Here we apply similar ideas to \cite{2021_EAAI_BeCAPTCHA_Acien} considering in this case mouse dynamics instead of touchscreen gestures, a richer signal in terms of time resolution, naturalness, and neuromotor information \cite{Plamondon}}.

Our contributions with this work go a step forward in the bot detection field for mouse dynamics, incorporating behavioral modeling and improved learning methods based on realistic synthetic samples (see Fig.~\ref{diagram}):

\begin{figure}
\noindent\makebox[\textwidth]{
\centering
\includegraphics[width=1.2\textwidth]{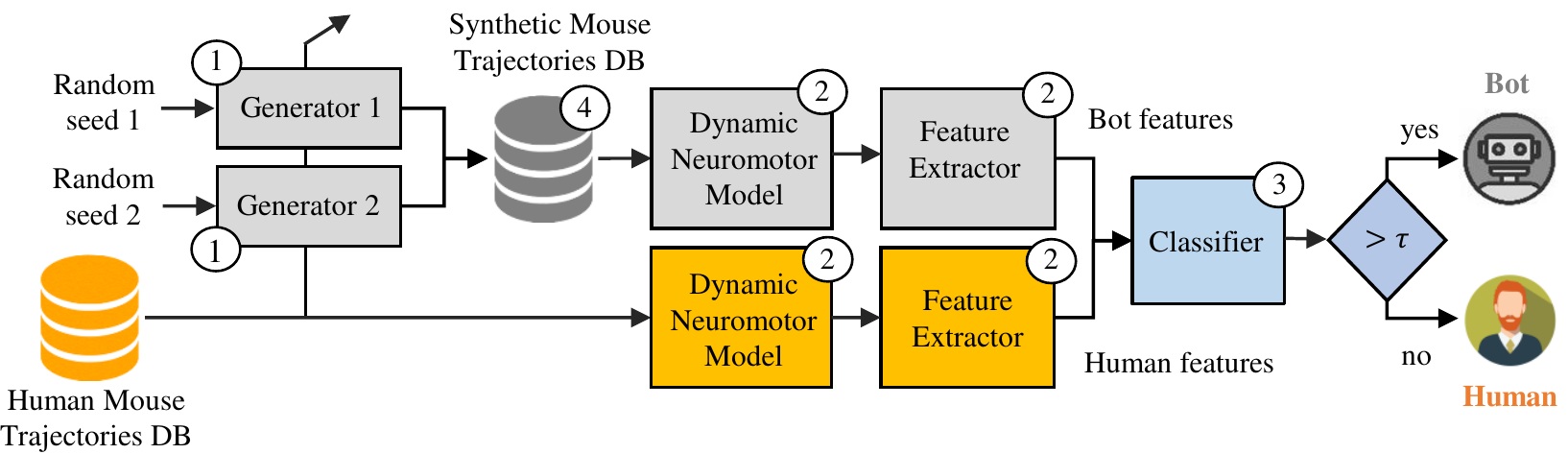}}
\caption{Learning framework of BeCAPTCHA-Mouse: \textcolor{black}{(1) We propose two novel methods to generate realistic synthetic mouse trajectories that allow to train and evaluate bot detection systems based on mouse dynamics; (2) We propose a neuromotor model to characterize Human and Synthetic Mouse Trajectories; (3) We evaluate the proposed features using multiple classifiers and learning scenarios; and (4) The proposed Generators can be also helpful for other HCI applications.}}
\label{diagram}
\end{figure}

\begin{itemize}
\setlength\itemsep{0em}
    \item (1) We propose two new methods for generating realistic mouse trajectories: \textit{i)} a Function-based method based on heuristic functions, and \textit{ii)} a data-driven method based on GANs in which a Generator synthesizes human-like trajectories from a Gaussian noise input. We demonstrate the usefulness of these synthetic trajectories to train more accurate bot detectors. These Generators can be helpful in many HCI research areas and applications.

    \item (2) We propose BeCAPTCHA-Mouse, a new bot detector based on neuromotor modeling \cite{Plamondon} of mouse trajectories and supervised classification trained with human and synthetic data. As showed in Fig. \ref{visual}, our proposed mouse detection algorithm can be added in a transparent setup and enhance traditional CAPTCHAs based on cognitive challenges, for example when you select the images in a visual CAPTCHA, or when you navigate through a website.

\begin{figure}[t!]
\centering
\includegraphics[width=1.0\columnwidth]{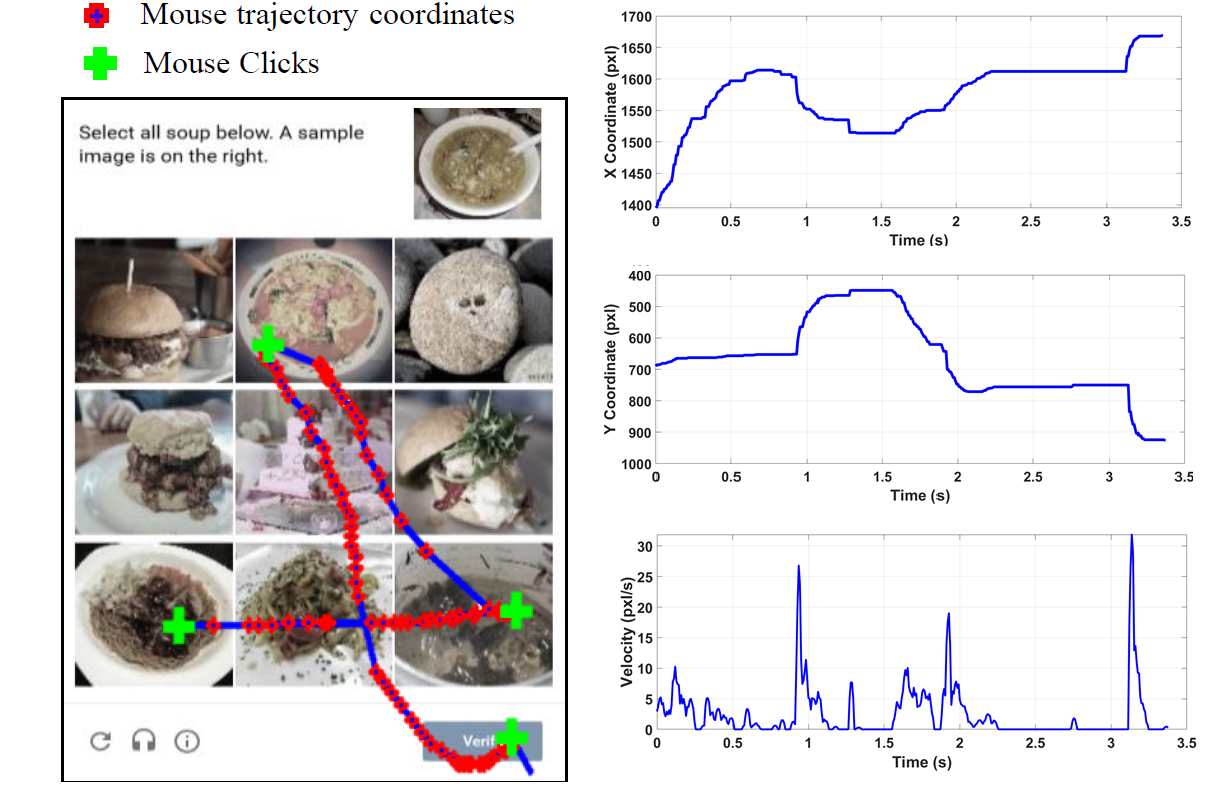}
\caption{An application example of our proposed mouse bot detection algorithm in combination with a traditional image-based CAPTCHA. While the user completes the image CAPTCHA task (cognitive challenge, left), our algorithm analyzes the mouse dynamics performed during the task (\{$\mathbf{x}$, $\mathbf {y}$\} coordinates and velocity profile, right).}
\label{visual}
\end{figure}
    
    \item \textcolor{black}{(3) Our experiments consider a large number of state-of-the-art classifiers and provide a detailed study, exposing the strengths and weakness of the classifiers in different scenarios. The experiments include: Support Vector Machine (SVM), Random Forest (RF), K-Nearest Neighbors (KNN), Multi-Layer Perceptron (MLP), and deep learning architectures such us Long Short-Term Memory (LSTM) and Gated Recurrent Units (GRUs). These algorithms are evaluated for mouse trajectories with different characteristics (e.g. direction, length) and learning strategies (e.g. number of samples, supervised, non supervised).}  
    
    \item (4) We present BeCAPTCHA-Mouse Benchmark\footnote{\hyperlink{https://github.com/BiDAlab/BeCAPTCHA-Mouse}{https://github.com/BiDAlab/BeCAPTCHA-Mouse}}, the first public benchmark for mouse-based bot detection including $10$,$000$ human and synthetic trajectories generated according to $10$ different types of synthesized behaviors. The inclusion of various types of synthetic samples (both for training and testing BeCAPTCHA-Mouse) allows to train strong bot detectors. Also, it allows comprehensive evaluations under various conditions including the worst-case scenario in which bot attacks mimic human behavior using latest machine learning advances. This benchmark can be helpful for other HCI applications involving mouse dynamics beyond bot detection.

\end{itemize}

\textcolor{black}{The main drawback of traditional CAPTCHA methods is that they only measure cognitive human skills (e.g. character recognition from distorted images, class-objects identification in a set of images, or speech translation from distorted audios). Trying to ensure a very accurate bot detection makes these CAPTCHAs difficult to perform even for humans. The main goal of our proposed method is to focus more on human behavioral skills rather than on cognitive ones. Neuromotor skills reveal human features useful for bot detection just with simple mouse trajectories. To the best of our knowledge, there are only a very limited number of works using mouse biometrics for bot detection. The most related to our research are \cite{Chu2018} and \cite{Ismail}. In \cite{Ismail} they synthetize mouse trajectories over a grid to hack the Google reCAPTCHA v3 algorithm, and in \cite{Chu2018} they extract global features (e.g. duration, average speed, displacement) from mouse and keystroke patterns to conduct a case study in the detection of blog bots for online blogging systems. While previous work in mouse dynamics (\cite{Ahmed, Chu2018}) focused on basic cues like duration or average speed, in this work we go a step forward by focusing on the analysis and synthesis of entire mouse trajectories. We propose to use the Sigma-Lognormal model to extract human features that characterizes better human behaviors and novel generation methods to synthesize human-like trajectories to improve the training and evaluation of these methods.}

The rest of the paper is organized as follows. In Section \ref{behave} we first discuss the usage of mouse dynamics in the context of behavioral biometrics. Section \ref{system_description} presents our bot detector BeCAPTCHA-Mouse. \textcolor{black}{Section \ref{neuromotor_model} introduces the mouse dynamics neuromotor model and the features employed for the classification of bot and human trayectories. Section \ref{sec:synthesis} describes the proposed methods for generating synthetic mouse trajectories.} Section \ref{results} describes our experimental framework (BeCAPTCHA-Mouse Benchmark) and presents the results obtained. Finally, Section \ref{conclusions} summarizes the conclusions and future works.
\section{Mouse Dynamics in the Context of Behavioral Biometrics}
\label{behave}

Human-Machine interaction generates a heterogeneous flow of data from multiple channels. This interaction generates patterns affected by: humans (e.g. attitude, emotional state, neuromotor, and cognitive abilities), sensor characteristics (e.g. ergonomics, precision), and task characteristics (e.g. easy of use, design, usefulness). Modeling the user behavior using these heterogeneous data streams is an ongoing challenge with applications in a variety of fields such as security, e-health, gaming, or education \cite{picard2020behavior,Javier_2020,pentland2018biometrics}. Among this variety of data sources, in the present paper we concentrate in behavioral biometric signals \cite{kalita2016smartphone}.

The literature of behavioral biometrics in the context of Human-Computer Interaction is large and includes several characteristics, e.g.: keystroking \cite{2016_IEEEAccess_KBOC_Aythami,alej2020typenet}, handwriting \cite{2021_TBIOM_DeepSign_Tolosana}, touchscreen signals \cite{2018_TIFS_Swipe_Fierrez,COMPSAC_ACien}, stylometry \cite{Locklear}, and mouse dynamics \cite{Ahmed}. Each characteristic has its pros and cons, therefore, a single biometric characteristic is usually not suitable for all applications. The biometric research community has identified several factors that determine the suitability of a biometric characteristic to be used in a certain application \cite{Anil}.

\begin{table}[]
\centering
\resizebox{1.0\textwidth}{!}{
\begin{tabular}{l|c|c|c|c|c|c|c|c}
 & \textbf{Uniq.} & \textbf{Univ.} & \textbf{Meas.} & \textbf{Perf.} & \textbf{Circ.} & \textbf{Acce.} & \textbf{Cog.} & \textbf{Neu.} \\ \hline\hline
Keystroke  & ** & ** & *** & *** & ** & **  & **  & *** \\
Stylometry & *  & *  & *   & *   & *  & *   & *** & *   \\
Web-log      &  ** & * & ***& ** & *  & * & ***  & * \\
\textbf{Mouse}   & * & **  & ***  & ***  & *  & ***   & ** & ***  
\end{tabular}%
}
\caption{Biometric characteristics typically obtained in human-computer interaction. We rate each factor with * (low), ** (medium), and *** (high). Uniq = Uniqueness, Univ = Universality, Meas = Measurability, Perf = Performance, Circ = Circumvention, Acce = Acceptability, Cog = Cognitive, Neu = Neuromotor.}
\label{Factors}
\end{table}

Table \ref{Factors} rates these factors for biometrics characteristics typically obtained from Human-Computer Interaction highlighting Mouse Dynamics, the focus in the present paper. Note that we added two factors related to the nature of the patterns obtained from these characteristics (Cognitive and Neuromotor patterns) with respect to the characteristics defined by \cite{Anil}.  

Now focusing in mouse dynamics for biometrics, in \cite{Ahmed} researchers explored characteristics obtained from mouse tasks for user recognition. They analyzed $68$ global features (e.g. duration, curvature, mean velocity) from mouse dynamics extracted during login sessions. Their results achieve up to $95\%$ authentication accuracy for passwords with $15$ digits. Besides, mouse dynamics can be combined with keystroke biometrics for continuous authentication schemes \cite{sim2007continuous}. The fusion of both biometric modalities has been shown to outperform significantly each individual modality achieving up to $98\%$ authentication accuracy \cite{Mondal}. In \cite{Albo}, the authors applied the Sigma-Lognormal Model based on the Kinematic Theory \cite{Plamondon} to compress mouse trajectories. They suggested that mouse movements are the result of complex human motor control behaviors that can be decomposed in a sum of primal movements. In addition, in \cite{Chen}, the authors studied the relationship between eye gaze position and mouse cursor position on a computer screen during web browsing and suggested that there are regular patterns of eye/mouse movements associated to the motor cortex system.

\section{BeCAPTCHA-Mouse: Bot Detection based on Mouse Dynamics}
\label{system_description}

The mouse is a very common device and its usage is ubiquitous in human-computer interfaces. Bot detection based on mouse dynamics can be therefore applied either in active or passive detectors.

Our BeCAPTCHA-Mouse bot detector \textcolor{black}{is based on two main pillars: 1)} we use mouse dynamics to extract neuromotor features capable to distinguish human behavior from bots (see Fig. \ref{diagram}); \textcolor{black}{2) we generate synthetic mouse trajectories to improve the learning framework of bot detectors.}

Mouse dynamics are rich in patterns capable of describing neuromotor capacities of the users. Note that we do not claim to replace other approaches (e.g. Google's reCAPTCHA) by mouse-based bot detection, our purpose is to enhance them by exploiting the ancillary information provided by mouse dynamics (see Fig.~\ref{visual}).

Our proposed method for bot detection consists in characterizing each mouse trajectory \textcolor{black}{(real and synthetic)} with a fixed-size feature vector \textcolor{black}{obtained from a neuromotor decomposition of the velocity profile}, followed by a standard classifier. Each trajectory characterized in this way can be classified individually using standard classifiers into human or bot based on supervised training using a development groundtruth dataset. When multiple trajectories are available, standard information fusion techniques can be applied \cite{2018_INFFUS_MCSreview1_Fierrez}. The more realistic the synthetic data used as groundtruth for training the classifier the stronger the classifier. 

In our experimental work we \textcolor{black}{demonstrate the effectiveness of the neuromotor features and the synthetic samples for different} classifiers. The contribution and success of our BeCAPTCHA-Mouse bot detector is not in the particular classifier used, but in two other fronts (see Fig.~\ref{diagram}): the high realism of the groundtruth data used for training our classifiers (with the methods presented in Section \ref{sec:synthesis}), and our proposed trajectory modeling using neuromotor features.

\subsection{BeCAPTCHA-Mouse: Neuromotor Analysis of Mouse Trajectories}
\label{neuromotor_model}

\begin{figure}[t!]
\noindent\makebox[\textwidth]{
\centering
\includegraphics[width=1.2\textwidth]{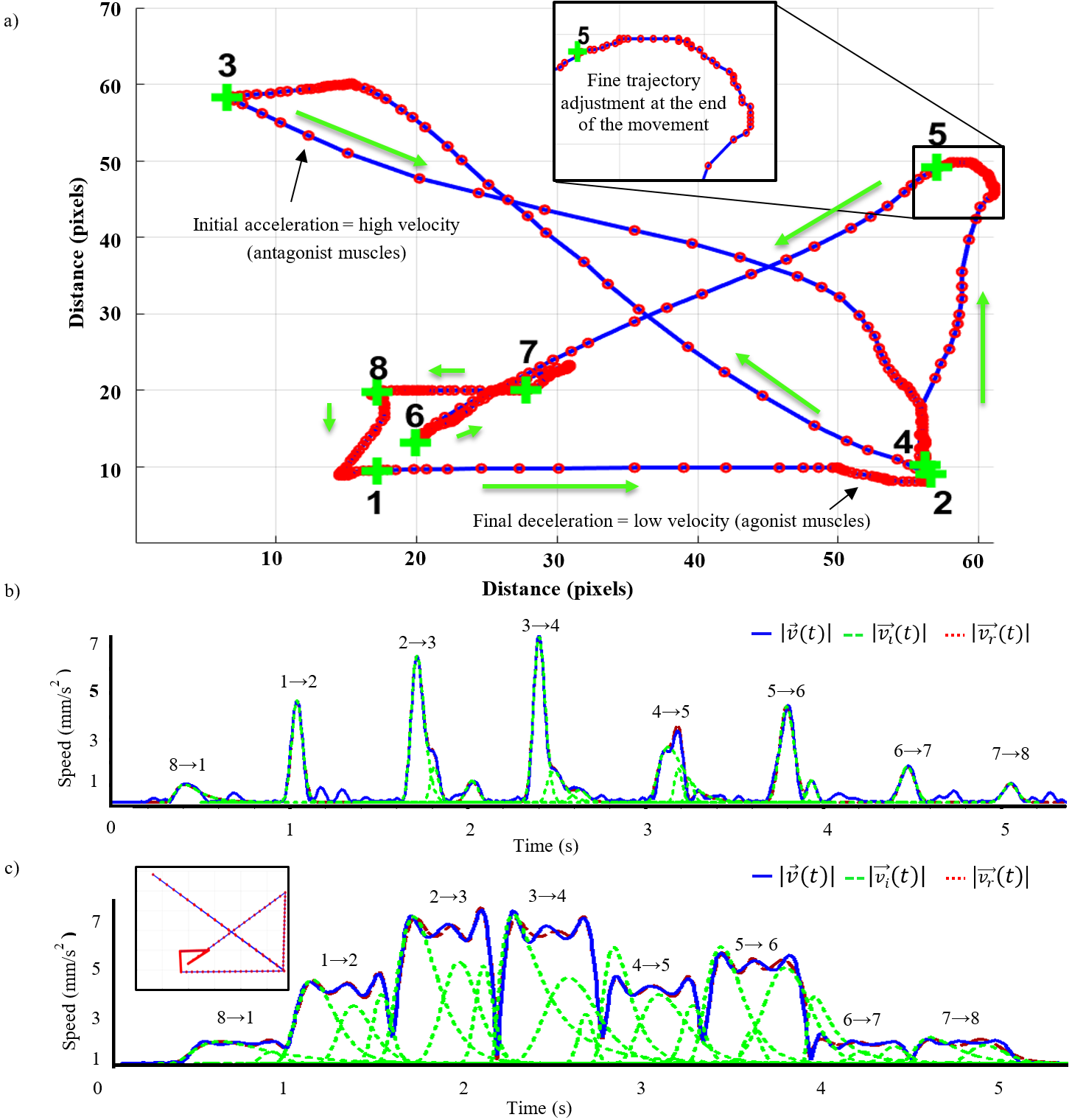}}
\caption{a) Example of the mouse task determined by $8$ keypoints:  the crosses represent the keypoint where the user must click, red circles are the ($x$,$y$) coordinates obtained from the mouse device, and the black line is the mouse trajectory. b) and c) are examples of the Lognormal decomposition of a human mouse movement and a synthetic linear trajectory respectively. }
\label{Fig:database}
\end{figure}

By looking at typical mouse movements (see Fig. \ref{Fig:database}.a), we can observe some aspects typically performed by humans during mouse trajectories execution: an initial acceleration and final deceleration performed by the antagonist (activate the movement) and agonist muscles (opposing joint torque) \cite{Plamondon}, and a fine-correction in the direction at the end of the trajectory when the mouse cursor gets close to the click button (characterized by a low velocity that serves to improve the precision of the movement). These aspects motivated us to use neuromotor analysis to find distinctive features in human mouse movements. Neuromotor-fine skills, that are unique of human beings are difficult to emulate for bots and could provide distinctive features in order to tell humans and bots apart. 

For this, we propose to model the trajectories according to the Sigma-Lognormal model \cite{Fischer} from the kinematic theory of rapid human movements \cite{Plamondon}. The model states that the velocity profile of the human hand movements (mouse movements in this work) can be decomposed into primitive strokes with a Lognormal shape that describes well the nature of the hand movements ruled by the motor cortex. The velocity profile of these strokes is modeled as:

\begin{equation}
\label{velocity_stroke}
     \left | \vec{v_{i}} \left ( t \right )\right |= \frac{D_{i}}{\sqrt{2\pi}\sigma_{i}\left ( t-t_{0i} \right )}\exp\left ( \frac{\left ( \ln\left ( t-t_{0i} \right ) -\mu_{i} \right )^{2}}{-2\sigma _{i}^{2}} \right )
\end{equation}

\noindent where the parameters are described in Table \ref{Table:Neuromotor_parameters}. The velocity profile of the entire hand movement is calculated as the sum of all these individual strokes:
\begin{equation}
\label{velocity}
     \vec{v_{r}} \left ( t \right )= \sum_{i=1}^{N}\vec{v_{i}} \left ( t \right )
\end{equation}

\begin{table}[]
\centering
\resizebox{0.75\textwidth}{!}{
\begin{tabular}{l|l}
\textbf{Parameter} & \textbf{Description}                                 \\ \hline\hline
$D_{i}$                & Input pulse: covered distance \\
$t_{0i} $              & Initialization time: displacement in the time axis   \\
$\mu_{i}  $              & Log-temporal delay                                   \\
$\sigma_{i}  $              & Impulse response time of the neuromotor system       \\
$\theta_{si} $              & Starting angle of the stroke                         \\
$\theta_{ei}  $            & Ending angle of the stroke                          
\end{tabular}%
}
\caption{Sigma-Lognormal features description.}
\label{Table:Neuromotor_parameters}
\end{table}

\noindent where $N$ is the number of velocity strokes considered in the model. A complex action like handwriting signature or mouse movements, is a summation of these lognormals, each one characterized by the six parameters in Table \ref{Table:Neuromotor_parameters}. An example of this is shown in Fig. \ref{Fig:database}.b, where the black line is the velocity profile $\left | \vec{v} \left ( t \right ) \right |$   of the above human mouse task (Fig. \ref{Fig:database}.a), which is used as the input of the Sigma-Lognormal model. The green dashed lines correspond to the individual lognormal signals $ \left |\vec{v_{i}} \left ( t \right )\right |$  generated as in \cite{Fischer}, which describes a method to automatically estimate both $N$ and the parameters in Table \ref{Table:Neuromotor_parameters} from an input trajectory $\left | \vec{v} \left ( t \right ) \right |$. Finally, the red dotted line  $\left |\vec{v_{r}}\left(t\right)\right|$  is the reconstruction of the original velocity profile by summing all these generated individual lognormal signals. We can observe that the reconstructed signal matches almost perfectly with the original velocity profile of the human mouse movement, suggesting the potential of the Sigma-Lognormal model to describe neuromotor mouse movements. Lognormals with a high amplitude are typically observed during the first part of the movement (agonist and antagonist activations), while smaller lognormals occur during the fine correction. The differences in lognormal sizes provide us information about the length of the trajectory (long trajectories have usually larger velocities).

The neuromotor feature set proposed for bot detection is computed from the six lognormal parameters described in Table \ref{Table:Neuromotor_parameters}. Each mouse trajectory generates $N$ lognormal signals and each lognormal generates those $6$ parameters from Table \ref{Table:Neuromotor_parameters}. For each parameter, we calculate $6$ features: maximum, minimum, and mean for both halves of the trajectory. This is done because in natural mouse movements the lognormal parameters are usually very different between both halves of a given trajectory (e.g. Fig. \ref{Fig:database}.b). Additionally, we added the number of lognormals $N$ that each mouse trajectory generates as an additional feature. This additional feature measures the complexity of the trajectory \cite{Lognormality_sign}, having many lognormals means that the mouse trajectory has many changes in the velocity profile while few of them usually indicates more basic trajectories. As a result, the neuromotor feature set has size $37$.

\subsection{BeCAPTCHA-Mouse: Trajectory Synthesis}
\label{sec:synthesis}

In the present paper, a mouse movement is defined by the spatial trajectory across time between two consecutive clicks, i.e., a sequence of points \{$\mathbf{x}$, $\mathbf {y}$\} \textcolor{black}{and a velocity profile $\left | \vec{v} \left ( t \right ) \right |$}, where $\mathbf{x}=[x_1,\ldots,x_M$], $\mathbf{y}=[y_1,\ldots,y_M$], and $M$ is the number of time samples. A mouse trajectory is defined by two main characteristics: the shape \textcolor{black}{(defined by \{$\mathbf{x}$, $\mathbf {y}$\})} and the velocity profile \textcolor{black}{(defined by $\left | \vec{v} \left ( t \right ) \right |$)}. In order to generate realistic synthetic samples, both characteristics must be considered in the generation method. We propose two methods for synthetically generating such mouse movement

\subsubsection{Method 1: Function-based Trajectories}

We generate mouse trajectories according to three different trajectory shapes (linear, quadratic, and exponential) and three different velocity profiles (constant, logarithmic, and Gaussian). 
We can synthesize many different mouse trajectories that mimic human movements by varying the parameters of each function. To generate a synthetic trajectory \{$\mathbf{\hat{x}}$, $\mathbf {\hat{y}}$\} with $M$ points, first we define the initial point $[\hat{x}_1$, $\hat{y}_1]$ and ending point $[\hat{x}_M$, $\hat{y}_M]$. Second, we select one of three velocity profiles \textcolor{black}{$\left | \vec{\hat{v}} \left ( t \right ) \right |$:} \textit{i)} constant velocity, where the distance between adjacent points is constant; \textit{ii)} logarithmic velocity, where the distances are gradually increasing (acceleration); and \textit{iii)} Gaussian velocity, in which the distances first increase and then decrease when they get close to the end of the trajectory (acceleration and deceleration). Third, we generate a sequence $\mathbf{\hat{x}}$ between $\hat{x}_1$ and $\hat{x}_{M}$ spaced according to the selected velocity profile. The $\mathbf{\hat{y}}$ sequence is then generated according to the shape function. For example, for a shape defined by the quadratic function $\hat{y} = a\hat{x}^{2} + b\hat{x} + c$, we fit $b$ and $c$ for a fixed value of $a$ by using the initial and ending points. We repeat the process fixing either $b$ or $c$.  The range of the parameters \{$a$, $b$, $c$\} explored is determined by analyzing real mouse movements fitted to quadratic functions. Linear and exponential shapes are generated similarly.

\begin{figure*}[t!]
\noindent\makebox[\textwidth]{
\centering
\includegraphics[width=1.1\textwidth]{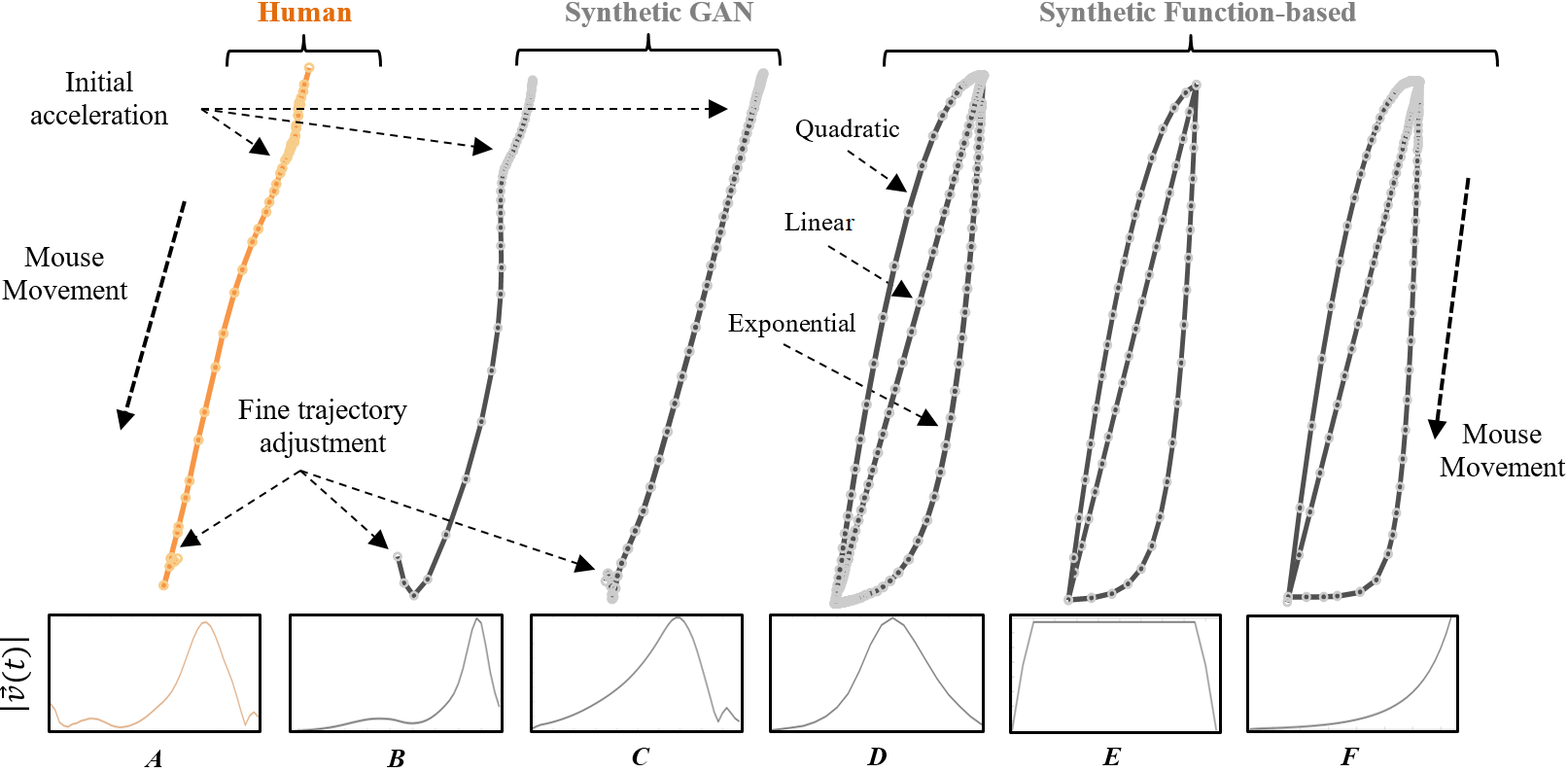}}
\caption{Examples of mouse trajectories and their velocity profiles employed in this work: $A$ is a real one extracted from a task of the database; $B$ and $C$ are synthetic trajectories generated with the GAN network; $D$, $E$ and $F$ are generated with the Function-based approach. Note that for each velocity profile ($D$ = Gaussian, $E$ = constant, $F$ = logarithmic), we include the three Function-based trajectories (linear, quadratic, and exponential).}
\label{Fig:trajectories}
\end{figure*}

Fig.~\ref{Fig:trajectories} (trajectories $D$, $E$, and $F$) shows some examples of these mouse trajectories synthesized. That figure also shows the $3$ different velocity profiles considered: the $3$ trajectories in $E$ have constant velocity, $F$ shows acceleration (the distance between adjacent samples increases gradually), and $D$ has initial acceleration and final deceleration. We can generate infinite mouse trajectories with this approach by varying the parameters of each function.

An important factor when synthetizing mouse trajectories is the number of points ($M$) of the trajectory. This usually varies depending not only on the length of the trajectory, but also on the direction, because different muscles are involved when we perform mouse trajectories in different directions. To emulate this phenomenon, we calculate the mean and standard deviation of the number of points for each of the $8$ mouse trajectories from the human data used in the experiments. Then, we synthetize trajectories with different number of points following a Gaussian distribution with the calculated mean and standard deviation.

\subsubsection{Method 2: GAN-based Trajectories}

\begin{figure*}[t!]
\noindent\makebox[\textwidth]{
\centering
\includegraphics[width=1.1\textwidth]{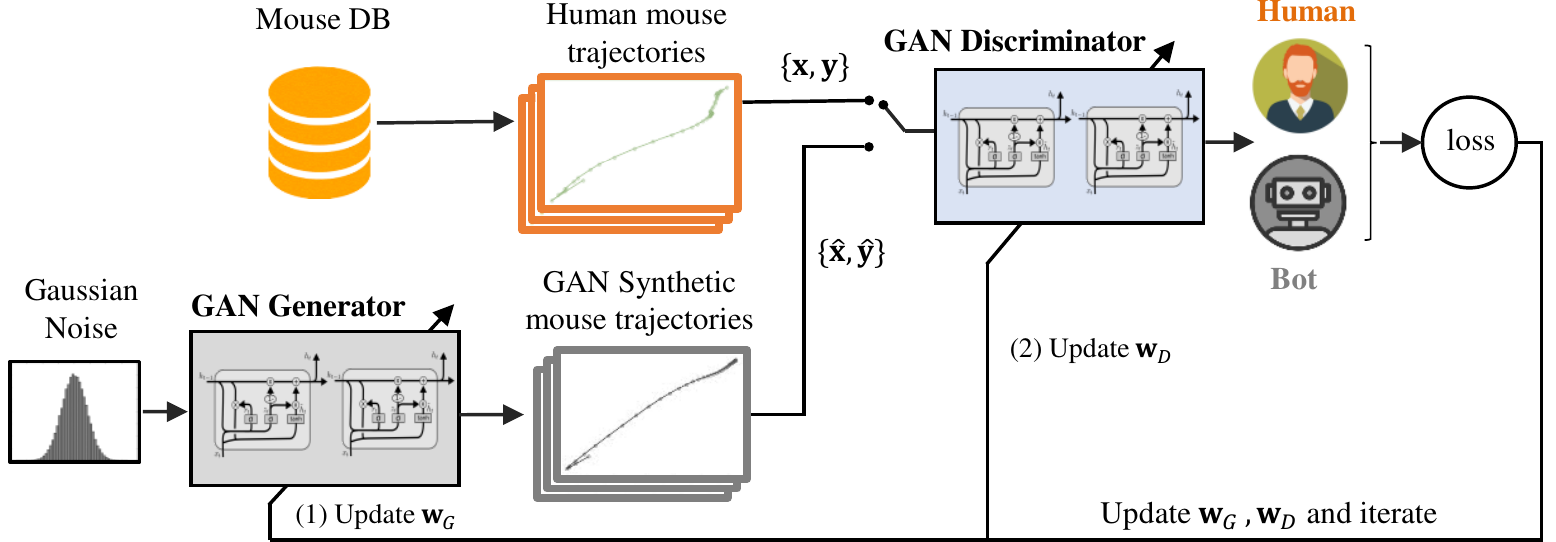}}
\caption{The proposed architecture to train a GAN Generator of synthetic mouse trajectories. The Generator learns the human features of the mouse trajectories and generate human-like ones from Gaussian Noise. \textcolor{black}{Note that the weights of the Discriminator $\mathbf{w}_D$ are trained after the update of the weights of the Generator $\mathbf{w}_G$.}}
\label{Fig:GAN}
\end{figure*}

For this approach we employ a GAN (Generative Adversarial Network) \cite{2020_JSTSP_GANprintR_Neves}, in which two neuronal networks, commonly named Generator \textcolor{black}{(defined by its parameters $\mathbf{w}_G$)} and Discriminator \textcolor{black}{(defined by its parameters $\mathbf{w}_D$)}, are trained one against the other (thus the “adversarial”). The architecture of the GAN is depicted in Fig.~\ref{Fig:GAN}. The aim of the Generator is to fool the Discriminator by generating fake trajectories \textcolor{black}{\{$\hat{\mathbf{x}}$, $\hat{\mathbf{y}}$\} very similar to the real ones \{$\mathbf{x}$, $\mathbf {y}$\}. We used a fixed sampling rate of $200$Hz for all the real and generated trajectories. The sampling rate is determined by the real trajectories used in the learning framework ($200$Hz in our experiments). Therefore, the synthesized samples are generated with the same sampling rate. Other frequencies can be obtained subsampling the generated ones or re-training the GAN for a different sampling rate. The input of the Generator consist of a seed vector of $R$ random numbers (in our experiments $R=100$). The output of the Generator are two coordinate vectors \{$\hat{\mathbf{x}}$, $\hat{\mathbf{y}}$\} with length equal to $M$ ($M$ can be fixed to generate different lengths). The input of the Discriminator consists of a batch including two types of trajectories: 1) \textit{Bot}: synthetic trajectories generated by the Generator; 2) \textit{Human}: real mouse trajectories chosen randomly from the Mouse DB described in next sections. The aim of the Discriminator is to predict whether the sample comes from the human set or is a fake created by the Generator. During the training phase, the GAN architecture will improve the ability of the Generator to fool the Discriminator. This architecture turns latent space points defined by the random seed into a classification decision: ‘\textit{Bot}’ (from the Generator) or ‘\textit{Human}’. This learning process is guided by the real mouse trajectories from the Mouse DB. During the GAN training, the weights of the Discriminator ($\mathbf{w}_D$) remain frozen. The iterative training process will update the weights $\mathbf{w}_G$ of the Generator in a way that makes Discriminator more likely to predict ‘\textit{Human}’ when looking at synthetic mouse trajectories. If the Discriminator is not frozen during this process, it will tend to predict ‘\textit{Human}’ for all samples. The Discriminator is trained (weights $\mathbf{w}_D$ updated) after the update of the  weights of the Discriminator ($\mathbf{w}_G$). This process is repeated iteratively ($50$ epochs in our experiments). Once the Generator is trained this way, then we can use it to synthesize mouse trajectories very similar to the human ones.}   

\textcolor{black}{The topology employed in the Discriminator consist of two LSTM (Long Short-Term Memory) layers (with $128$ and $64$ units respectively, with ‘\textit{LeakyReLU}’ activation) followed by a dense layer (with $1$ unit and ‘\textit{Sigmoid}’ activation). The dense layer of the Discriminator is used as a classification layer to distinguish between bot and real mouse trajectories (‘\textit{Binary Cross-Entropy}’ loss function). For the Generator, we employ two LSTM layers (with $128$ and $64$ units respectively, with ‘\textit{ReLU}’ activation) followed by a dense layer with $2$ units (one unit for build each \{$\hat{\mathbf{x}}$, $\hat{\mathbf{y}}$\} mouse coordinates) and ‘\textit{TimeDistributed}’ activation.}

The GAN network was trained using $60\%$ of the human mouse trajectories in the database. Training details: learning rate $\alpha = 2 \times 10^{-4}$, Adam optimizer with $\beta_{1} = 0.5$, $\beta_{2} = 0.999$, $\epsilon = 10^{-8}$, $50$ epochs with a batch size of $128$ samples for both Generator and Discriminator.

Fig.~\ref{Fig:trajectories} shows two examples (trajectories $B$ and $C$) of synthetic mouse trajectories generated with the GAN network and the comparison with a real one. We can observe high similarity between the two synthetic examples and the real one. Human mouse patterns such us the initial acceleration and the final trajectory fine correction that we discussed before are automatically learned by the GAN network and reproduced in the synthetic trajectories generated.

\section{Experiments}
\label{results}

\subsection{BeCAPTCHA-Mouse Benchmark: Database}

The human mouse trajectories employed in this work were extracted from Shen \textit{et al.} database \cite{SHEN}, which is comprised of more than $200$K mouse trajectories acquired from $58$ users who completed 300 repetitions of the task. Acquisition of data from each subject took between 30 days and 90 days. In each repetition, the task was to click $8$ buttons that appeared in the screen sequentially. This task was repeated twice in each session. Fig. \ref{Fig:database}.a shows an example of the whole mouse movement task. Note that the buttons are placed in a particular order to generate mouse trajectories with different directions (rightwards, upwards, downwards, and oblique) and different lengths. 

In the present work, we define a mouse trajectory as the mouse displacement that occurs between two click buttons. Therefore, the mouse movement task of Fig. \ref{Fig:database}.a is composed of $8$ mouse trajectories. The raw data recorded during the acquisition process was: the mouse position over the screen (\{$\mathbf{x}$, $\mathbf {y}$\} axis position in pixels), the event (movement or click), and timestamp of the event. The experiments presented in this work are performed using a subset of the database including $35$ samples (randomly chosen) from each of the $58$ users available (more than $5$K trajectories in total).

Fig. \ref{Fig:database}.c shows the decomposition of a synthetic function-based trajectory with linear shape. We can observe the huge differences between both lognormal decompositions (the human trajectory and the synthetic one) by looking at the shape of the lognormal signals. The synthetic trajectory has wider lognormals and they are more symmetric than the human ones. Note that the Sigma-Lognormal algorithm introduces a low-pass filter to the input signal, that is the reason why the velocity profile of the synthetic trajectory (Fig. \ref{Fig:database}.c) is a bit smoothed, but the difference between both synthetic and human velocity profiles is still patent.

\textcolor{black}{The BeCAPTCHA-Mouse Benchmark is composed of $5$K human trajectories and $10$K synthetic trajectories generated according to the two methods proposed ($5$K function-Based and $5$K GAN trajectories). Both real and synthesized samples are characterized by a variety of lengths, directions, and velocities.}

\subsection{BeCAPTCHA-Mouse: Role of the Direction and Length of the Trajectory} \label{ExpProtocol}

We have extracted the proposed neuromotor features from human and synthetic mouse trajectories. \textcolor{black}{For this first experiment, we use a Random Forest (RF) classifier because of its best performance among all classifiers evaluated (as we will see in the next section).} The experiments are divided according to the $8$ real mouse trajectories present in the whole task. This means that we classify at trajectory level (i.e. the mouse trajectory performed between two consecutive click buttons) instead of classifying the whole task. This is because the task was designed to take into account trajectories with different directions and lengths, and therefore, different muscles configurations are involved in each trajectory. In this way, we can analyze which mouse trajectories are better to discriminate between humans and bots. \textcolor{black}{We train $10$ different RFs (one for each type of attack, see columns in Table~\ref{Resultados}) using both human and synthetic trajectories. For each RF, we train the classifier by using $70\%$ of all samples (up to $1$,$500$ samples available for each type of trajectory between both synthetic and real ones) randomly chosen as the training set. The other $30\%$ samples are employed for evaluation. The results are obtained by repeating each experiment $5$ times and averaging, with a standard deviation of $\sigma \sim 0.1\%$.}

\begin{table}[]
\arrayrulewidth=1pt
\resizebox{\textwidth}{!}{%
\noindent\makebox[1.2\textwidth]{
\begin{tabular}{|l |c |c|c|c|c|c|c|c|c|c|c|}
\hline
\multicolumn{2}{|c|} {} &
  \multicolumn{9}{c|}{Bot: Function-based} &
   \\ \cline{3-11}\cline{3-11} 
\multicolumn{2}{|c|}{} &
  \multicolumn{3}{c|}{Linear} &
  \multicolumn{3}{c|}{Quadratic} &
  \multicolumn{3}{c|}{Logarithmic} &
  \\ \cline{3-11}
\multicolumn{2}{|c|}{\multirow{-3}{*}{Trajectories}} &
  VP = 1 &
  VP = 2 &
  VP = 3 &
  VP = 1 &
  VP = 2 &
  VP = 3 &
  VP = 1 &
  VP = 2 &
  VP = 3 &
  \multirow{-3}{*}{\begin{tabular}[c]{@{}c@{}}Bot:\\ GAN\end{tabular}} \\ \hline
 &
  $8 \rightarrow 1$ &
  $98.6$ &
  $96.3$ &
  $99.0$ &
  $91.0$ &
  $91.0$ &
  $92.3$ &
  $89.0$ &
  $88.6$ &
  $89.3$ &
  $96.9$ \\ \cline{2-12}
 &
  $1 \rightarrow 2$ &
  $99.7$ &
  $98.6$ &
  $97.2$ &
  $91.6$ &
  $98.3$ &
  $92.2$ &
  $95.8$ &
  $92.3$ &
  $92.5$ &
  $96.7$ \\ \cline{2-12} 
 &
  $2 \rightarrow 3$ &
  $99.4$ &
  $99.1$ &
  $99.7$ &
  $95.3$ &
  $96.4$ &
  $88.0$ &
  $94.4$ &
  $98.9$ &
  $90.5$ &
  $99.9$ \\ \cline{2-12} 
 &
  $3 \rightarrow 4$ &
  $99.7$ &
  $97.5$ &
  $97.0$ &
  $94.2$ &
  $96.6$ &
  $90.5$ &
  $94.2$ &
  $95.1$ &
  $93.0$ &
  $99.7$ \\ \cline{2-12} 
 &
  $4 \rightarrow 5$ &
  $99.9$ &
  $98.0$ &
  $99.4$ &
  $95.5$ &
  $94.7$ &
  $92.5$ &
  $93.9$ &
  $95.4$ &
  $93.9$ &
  $97.0$ \\ \cline{2-12} 
 &
  $5 \rightarrow 6$ &
  $99.9$ &
  $98.9$ &
  $99.1$ &
  $92.8$ &
  $97.5$ &
  $91.4$ &
  $93.3$ &
  $95.1$ &
  $94.4$ &
  $98.3$ \\ \cline{2-12} 
 &
  $6 \rightarrow 7$ &
  $99.1$ &
  $98.3$ &
  $98.6$ &
  $90.2$ &
  $89.7$ &
  $93.6$ &
  $88.8$ &
  $92.3$ &
  $93.6$ &
  $98.1$ \\ \cline{2-12} 
\multirow{-8}{*}{\rotatebox{90}{Individual trajectories}} &
  $7 \rightarrow 8$ &
  $97.0$ &
  $96.6$ &
  $97.5$ &
  $92.2$ &
  $93.3$ &
  $93.0$ &
  $88.3$ &
  $88.6$ &
  $93.1$ &
  $98.7$ \\ \hline\hline
 &
  Neuromotor &
  $99.1$ &
  $98.7$ &
  $99.3$ &
  $96.9$ &
  $96.3$ &
  $94.7$ &
  $96.3$ &
  $95.2$ &
  $94.7$ &
  $98.0$ \\ \cline{2-12} 
 &
  Global Features {}\cite{Chu2018}{} &
   $99.7$ &
   $99.6$ &
   $99.7$ &
  $95.3$ &
  $96.7$ &
  $96.8$ &
  $97.2$ &
  $96.5$ &
  $97.3$ &
  $\textbf{99.8}$ \\ \cline{2-12} 
\multirow{-3}{*}{\rotatebox{90}{All}} &
  Neuromotor+\cite{Chu2018} &
  $\textbf{99.9}$ &
  $\textbf{99.7}$ &
  $\textbf{99.8}$ &
  $\textbf{98.0}$ &
  $\textbf{99.0}$ &
  $\textbf{98.4}$ &
  $\textbf{98.2}$ &
  $\textbf{98.9}$ &
  $\textbf{98.9}$ &
  $99.7$ \\ \hline
\end{tabular}}}
\caption{Accuracy rates ($\%$) in the binary classification between each of the 8 human trajectories and the synthetic ones. VP (Velocity Profile): VP = 1 constant velocity, VP = 2 initial acceleration, VP = 3 initial acceleration and final deceleration.}
\label{Resultados}
\end{table}

Table \ref{Resultados} shows the results for all classification schemes. The first $8$ rows present the $8$ trajectories derived from the movements between the $8$ keypoints (plotted in Fig. \ref{Fig:database}.a). The table shows the classification accuracy in $\%$ (human vs bot) for the different synthetic trajectories (in columns) generated in this work.

First, comparing among the different trajectories, we can observe that the shorter ones ($8\rightarrow1$, $6\rightarrow7$, and $7\rightarrow8$) show higher classification errors compared to the larger ones.  Short trajectories generate less neuromotor information: initial acceleration, final deceleration, and trajectory corrections are less pronounced in short trajectories. Second, logarithmic trajectory shapes achieve the worst classification performance, as we expected, because the shape of logarithmic functions fit better the human trajectories shapes. Third, the most significant parameter when synthetizing trajectories is the velocity profile. When VP = 3 (i.e., initial acceleration and final deceleration), the synthetic trajectories are able to fool the classifier up to $17\%$ of the times. This confirms that the velocity profile of human mouse trajectories plays and important role when describing human features in mouse dynamics. Four, the GAN Generator (last column in Table \ref{Resultados}) results in lower classification errors compared with the function-based method. This is surprising after visualizing the high similarity between human and GAN-generated trajectories (see Fig. \ref{Fig:trajectories} $A$ vs $B$ and $A$ vs $C$). We interpret this result with care: on the one hand it demonstrates that our bot detection approach is powerful against realistic and sophisticate fakes, but on the other hand the GAN Generator can be improved to better fool our detector. \textcolor{black}{Although the synthetic samples generated by the GAN Generator seems very realistic to the human eye, the RF classifiers were capable of detecting synthetic samples with high accuracy. These high classification rates suggest that GAN generators introduce patterns that allow its detection \cite{2020_JSTSP_GANprintR_Neves}.}

The last three rows in Table \ref{Resultados} present the results when features from all $8$ trajectories are combined (each RF is trained using features from all $8$ trajectories). Additionally, we compare the performance achieved with existing approaches \cite{Chu2018}. The feature set proposed in \cite{Chu2018} consists of $6$ global features: duration, distance, displacement, average angle, average velocity, and move efficiency (distance over displacement). The results suggest that the feature set proposed in \cite{Chu2018} outperforms the neuromotor features proposed here only for Linear synthetic trajectories. The best performance is obtained overall with an extended set composed by both sets of features. The extended set has the best results with an average around $99\%$ of accuracy independently of the type of synthetic trajectory.

\begin{table}[t!]
\normalsize
\centering
\begin{tabular}{l|c|c|}
\multirow{2}{*}{\textbf{Features} }    & \multicolumn{2}{c|}{\textbf{Training}} \\
 & Only Real \cite{Chu2018}  & Real+Fake {[}Ours{]}\\ \hline\hline
Global Features \cite{Chu2018}     &    $66.3\%$ $(\textrm{baseline})$      & $96.6\%$ $(\downarrow90.1\%)$               \\
Neuromotor {[}Ours{]}      &   $64.4\%$ $(\uparrow5.6\%)$  & $89.8\%$ $(\downarrow79.7\%)$              \\
Global+Neuromotor {[}Ours{]} & $59.9\%$ $(\uparrow19.0\%)$  & $98.2\%$ $(\downarrow95.4\%)$               
\end{tabular}%

\caption{Accuracy rates ($\%$) in bot detection of the different feature sets for models trained with and without synthetic samples (fakes) and evaluated using human samples and fake samples. One-Class SVM (first column) and Multiclass SVM (second columm). Relative error reduction with respect to the baseline \cite{Chu2018} in brackets.}
\label{Table:overall}
\end{table}

\subsection{BeCAPTCHA-Mouse: Role of Synthetic Samples}

Table \ref{Table:overall} shows the accuracy when all types of attacks are used to train and test the system. In this case, the classifier is trained using trajectories from all $8$ directions and synthetic samples from all $10$ types of attacks. \textcolor{black}{The Table shows the impact of introducing the synthetic samples (i.e. Real+Fake) in the learning process. For this experiment, we decided to use as classifiers a One-Class SVM (trained using only real trajectories) and a Multiclass SVM (trained using real and synthetic trajectories). The aim of the experiment is to evaluate to what extent the inclusion of synthetic samples in the learning framework serves to improve the accuracy of the model. The results show that the synthetic samples and neuromotor feature set proposed in this work allows to reduce the error by $95.4\%$ in comparison with the previous existing method \cite{Chu2018}. These results demonstrate the potential of synthetically generated trajectories and mouse dynamics features to boost the performance of new bot detection algorithms.} 

The results obtained show how training methods based on both real and synthetic trajectories clearly outperform training methods based exclusively on real samples. As can be seen, the classifier trained only with real samples was not capable to detect most of the attacks with accuracy rates lower than $70\%$ either for global features and neuromotor features. The importance of synthetic samples is twofold: i) evaluation of bot detection algorithms under challenging attacks generated according to different methods; and ii) training better detectors to model both human and synthetic behaviors. The results in Table \ref{Table:overall} show the potential of the synthetic samples and its usefulness to train better models capable to deal with all types of attacks.

\textcolor{black}{\subsection{BeCAPTCHA-Mouse: Ablation Study}}
\textcolor{black}{In this section we perform an ablation study on different classifiers to analyze their performance in bot detection for the $3$ multi-class scenarios proposed, according to the synthetic samples employed to train  and test them: Function-Based, GAN, and their Combination. It is worth noting that all classifiers are trained using trajectories from all $8$ directions and synthetic samples from all $10$ types of attacks, as reported in Table \ref{Table:overall} to allow fair comparisons.} 

\textcolor{black}{Table \ref{Table:Clasificadores} shows the performance of classification algorithms: Support Vector Machine (SVM) with a Radial Basis Function (RBF), K-Nearest Neighbors (KNN) with $ k = 10$, Random Forest (RF), Multi-Layer Perceptron (MLP), and $2$ Recurrent Neuronal Networks (RNN), (one composed by Long Short-Term Memory (LSTM) units and the other with Gated Recurrent Units (GRU). The RNNs (i.e. LSTM and GRU) were trained directly with the raw data (i.e. the sequence of points \{$\mathbf{x}$, $\mathbf {y}$\} of the mouse trajectories) instead of extracting the global features (i.e. Neuromotor + Baseline \cite{Chu2018}) as done with the statistical classifiers. The RNNs have the same architecture as the Discriminator of the GAN: two recurrent layers of $128$ and $64$ units respectively, followed by a dense layer to classify  between fake and real mouse trajectories. All classifiers were trained and tested following the same experimental protocol as in Section \ref{ExpProtocol}, using $70\%$ of all samples (up to $10$K samples between both real and synthetic samples when combining all types of trajectories) randomly chosen as the training set (named $L$ in this section, with $L= 7$,$000$). The results are reported in terms of Accuracy, AUC (Area Under the Curve), Precision, Recall, and F1.}

\arrayrulewidth=1pt
\begin{table}[]
\resizebox{\textwidth}{!}{%
\noindent\makebox[1.2\textwidth]{
\begin{tabular}{|l|l|l|l|l|l|l|l|l|l|l|l|l|l|l|l|}
\hline
\multirow{2}{*}{} & \multicolumn{15}{c|}{\textcolor{black}{Bot}}                                                                                   \\ \cline{2-16} 
                  & \multicolumn{5}{c|}{\textcolor{black}{Function-based}} & \multicolumn{5}{c|}{\textcolor{black}{GAN}}  & \multicolumn{5}{c|}{\textcolor{black}{Combination}} \\ \hline
\textcolor{black}{Classifiers}       & \textcolor{black}{Acc}   & \textcolor{black}{AUC}   & \textcolor{black}{Pre}   & \textcolor{black}{Re}    & \textcolor{black}{F1}   & \textcolor{black}{Acc}  & \textcolor{black}{AUC}  & \textcolor{black}{Pre}  & \textcolor{black}{Re}   & \textcolor{black}{F1}   & \textcolor{black}{Acc}    & \textcolor{black}{AUC}    & \textcolor{black}{Pre}    & \textcolor{black}{Re}    & \textcolor{black}{F1}    \\ \hline
\textcolor{black}{SVM}               & 98.0  & 99.4  & 98.6  & 96.7  & 97.7 & 98.5 & 99.6 & 99.2 & 97.9 & 98.5 & 98.2   & 99.4   & 97.3   & 99.0  & 97.4  \\ \hline
\textcolor{black}{KNN}               & 93.4  & 98.1  & 93.6  & 93.2  & 93.5 & 94.1 & 99.4 & 99.8 & 88.3 & 93.6 & 92.0   & 97.4   & 90.7   & 93.2  & 92.1  \\ \hline
\textcolor{black}{RF}                & \textbf{98.5}  & \textbf{99.8}  & \textbf{98.6}  & \textbf{98.8}  & \textbf{98.7} & \textbf{99.7} & \textbf{99.9} & \textbf{99.5} & \textbf{99.9} & \textbf{99.7} & \textbf{98.7}   & \textbf{99.9}   & \textbf{98.8}   & \textbf{99.0}  & \textbf{99.0}  \\ \hline
\textcolor{black}{MLP}   & 94.6  & 94.1  & 95.0  & 94.2  & 94.6 & 93.4 & 93.5 & 95.4 & 92.3 & 93.9 & 92.2   & 91.5   & 89.8   & 95.4  & 92.5  \\ \hline
\textcolor{black}{LSTM}   & 98.2  & 99.8  & 97.6  & 98.8  & 98.2 &  99.2 & 98.0 & 99.7 & 98.9 & 99.5  & 97.3   & 99.7   & 96.7  & 97.9 &  97.3\\ \hline
\textcolor{black}{GRU}   & 98.4 & 99.4  & 98.5  & 98.6 & 98.6 & 99.3 & 99.2 & 99.2 & 90.2 & 99.0 & 99.8   & 99.8   & 94.4   & 99.0  & 96.9  \\ \hline
\end{tabular}}}
\caption{Bot detection performance metrics in \% ( Acc = Accuracy, AUC = Area Under the Curve, Pre = Precision, Re = Recall, and F1) for the different scenarios: Function-based, GAN, and Combination.}
\label{Table:Clasificadores}
\end{table}

\textcolor{black}{First, we can observe that the best results among the statistical classifiers are achieved by the RF classifier followed by the SVM. KNN and MLP perform worst, although all classifiers have accuracy rates over $90\%$. Secondly,  among the different RNNs, the configuration with LSTM units performs sightly better than the one with GRU units, even though both recurrent network setups are outperformed by the RF classifier. These results suggest that the feature set chosen to train and test the statistical classifiers is suitable for the mouse bot detection task, outperforming other approaches based on deep neuronal networks architectures. Nonetheless, the RNNs demonstrate its capacity to extract useful features from the raw data.}

\textcolor{black}{In the next experiment we explore whether the number of training samples ($L$) plays and important role in the classification performance. We want to highlight that the training and the evaluation sets have the same number of human ($L_h$) and synthetic ($L_s$) samples, i.e.: $L_h = L_s = L/2$.}
\begin{figure*}[t!]
\noindent\makebox[\textwidth]{
\centering
\includegraphics[width=1.3\textwidth]{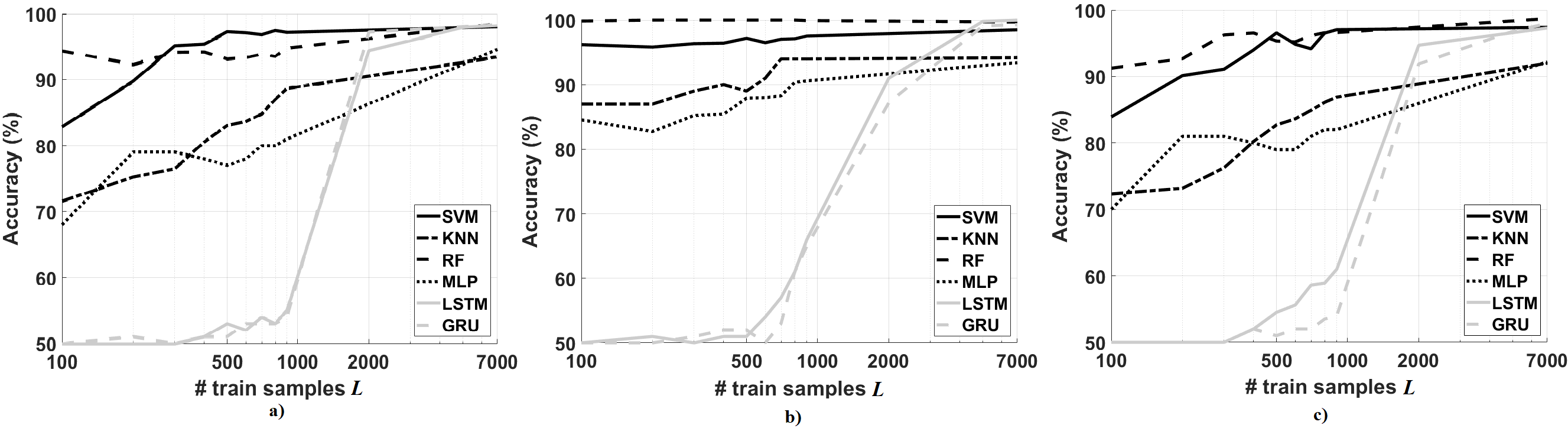}}
\caption{Accuracy curves ($\%$) against the number of train samples ($100 \leq L \leq 7$,$000$) to train the different classifiers in Function-based (a), GAN (b), and Combination (c) classification scenarios.}
\label{Fig:train_samples}
\end{figure*}

\textcolor{black}{For this, in Fig. \ref{Fig:train_samples} we plot the accuracy curves of the previous classifiers according to the number of samples employed in their training set. As expected, the accuracy improves in all scenarios when we enlarge the number of train samples. However, there are important differences between the statistical and the RNNs approaches. Meanwhile all statistical classifiers achieve their maximum performance with $L = 500$, both LSTM and GRU are not able to reach the same performance with only $500$ train samples. In fact, they need at least $L = 2$,$000$ to perform as well as the statistical classifiers. This shows the superior performance of the statistical classifiers in those scenarios where the number of samples to train the classifiers are scarce.}

\textcolor{black}{Finally, in the last experiment we replaced the previously introduced RNNs classifiers by the Discriminator model of the GAN architecture. The idea is to analyze in what extent the Discriminator of the GAN Network trained only with the synthetic samples generated by the Generator (and the real ones) during the GAN training could perform better in classification than the previous RNNs trained from scratch. For this, we tuned the number of neurons of the two LSTM layers of the Discriminator and trained a new GAN network for each Discriminator setup proposed.}

\arrayrulewidth=1pt
\begin{table}[]
\resizebox{\textwidth}{!}{%
\noindent\makebox[1.2\textwidth]{
\begin{tabular}{|l|l|l|l|l|l|l|l|l|l|l|l|l|l|l|l|}
\hline
\multirow{2}{*}{} & \multicolumn{15}{c|}{\textcolor{black}{Bot}}                                                                                   \\ \cline{2-16} 
                  & \multicolumn{5}{c|}{\textcolor{black}{Function-based}} & \multicolumn{5}{c|}{\textcolor{black}{GAN}}  & \multicolumn{5}{c|}{\textcolor{black}{Combination}} \\ \hline
\textcolor{black}{Discriminators}       & \textcolor{black}{Acc}   & \textcolor{black}{AUC}   & \textcolor{black}{Pre}   & \textcolor{black}{Re}    & \textcolor{black}{F1}   & \textcolor{black}{Acc}  & \textcolor{black}{AUC}  & \textcolor{black}{Pre}  & \textcolor{black}{Re}   & \textcolor{black}{F1}   & \textcolor{black}{Acc}    & \textcolor{black}{AUC}    & \textcolor{black}{Pre}    & \textcolor{black}{Re}    & \textcolor{black}{F1}    \\ \hline
\textcolor{black}{LSTM (128/64)}       & \textbf{89.9}  & \textbf{93.2}  & \textbf{88.5}  & \textbf{90.0} & \textbf{89.3} & 96.8 & 99.6 & 95.0 & 98.7 & 96.8 & \textbf{89.6}  & \textbf{93.9}   & \textbf{89.2}   & \textbf{90.0}  & \textbf{89.6}  \\ \hline
\textcolor{black}{LSTM (64/32)}       & 74.0  & 72.1  & 67.0  & 95.6  & 78.7 & \textbf{99.9} & \textbf{99.9} & \textbf{99.9} & \textbf{99.9} & \textbf{99.9} & 73.0   & 76.1  & 65.9   & 96.0  & 78.1  \\ \hline
\textcolor{black}{LSTM (32/16)}         & 81.4  & 80.2 & 77.9  & 88.0  & 82.6 & 99.7 & 98.9 & 99.6 & 99.9 & 99.8 & 78.8  &  76.0  & 74.4   & 88.0  & 80.6  \\ \hline
\textcolor{black}{LSTM (16/8)}           & 56.8 & 58.6 & 54.2  & 86.8 & 66.7  & 56.2 & 91.3 & 53.3 & 99.9 & 69.5 &  64.0  &  67.0  &  59.5  & 87.2  &  70.7 \\ \hline
\end{tabular}}}
\caption{Performance metrics in \% (AUC = Area Under the Curve, Acc, Pre, Re, and F1) for the different setups of GAN Discriminator in bot detection. In brackets the number of neurons for the first/second LSTM layer respectively used in the Discriminator.}
\label{Table:Discriminadores}
\end{table}

\textcolor{black}{Table \ref{Table:Discriminadores} shows the performance of $4$ GAN Discriminator setups for the $3$ classification scenarios proposed: the function-based, GAN, and their Combination. As we expected, the performance using GAN classification is much better than the performance achieved by the LSTM and GRU networks of the previous experiment, due to the Discriminators were trained specifically to discriminate between the synthetic mouse trajectories generated by the GAN Generator and the human ones. However, the Discriminators also classify quite well in the function-based scenario, even though no Function-based sample was employed to train them ($L_s = 0$). In fact, as we increase the complexity of the Discriminator with more neurons in both layers, the performance improves up to $90$\% of accuracy, close to the results achieved by the LSTM and GRU networks trained with $L_s = 7$,$000$ samples. These results show the potential of the GAN architecture, not only to generate synthetic mouse trajectories with similar shape to the human ones with the Generator, but also for classification purposes, as the Discriminator is able to classify between human and bot trajectories even against synthetic trajectories not seen during the training phase.}



\section{Conclusions and Future work}
\label{conclusions}
We have explored behavioral biometrics for bot detection during human-computer interaction. In particular, we have analyzed the capacity of mouse dynamics to describe human neuromotor features. Our conclusions in comparison to state-of-the-art works suggest that there is unexploited potential of mouse dynamics as a behavioral biometric for tasks such as bot detection.

In particular, we have proposed BeCAPTCHA-Mouse, a bot detection algorithm based on mouse dynamics, and a related benchmark\footnote{\hyperlink{https://github.com/BiDAlab/BeCAPTCHA-Mouse}{https://github.com/BiDAlab/BeCAPTCHA-Mouse}}, the first one public for research in bot detection and other mouse-based research areas including HCI, security, and human behavior.

\textcolor{black}{Our method is based on neuromotor features extracted from each mouse trajectory and a learning framework including both real and synthetic samples.} We have proposed and studied two new methods for generating synthetic mouse trajectories of varying level of realism. These generators are very useful both training stronger bot detectors and evaluating them in comprehensive and worst case scenarios. These generators are also valuable for related research problems beyond bot detection involving mouse dynamics.

In our experiments we have observed the main features of human mouse trajectories (e.g. initial acceleration, final deceleration, and fine trajectory correction). Based on that we have developed a neuromotor feature representation using the Sigma-Lognormal model \cite{Plamondon, Fischer}. Using the proposed neuromotor feature representation and training standard classifiers making use of the proposed synthetic mouse trajectories, we have been able to discriminate between humans and bots with up to $98.7\%$ of accuracy, even with bots of high realism, and only one mouse trajectory as input (between two consecutive clicks). This proves the potential of mouse dynamics for Turing tests. \textcolor{black}{Additionally, we also provided an exhaustive ablation study on different classifiers to explore the capacity of these algorithms for the bot detection task. Random Forests (RF) have demonstrated to perform the best in all scenarios evaluated followed by an LSTM network. However, when the number of train samples is reduced ($L\leq 1$,$000$), the LSTM is not able to classify as well as the RF classifier. In fact, the LSTM can be replaced by the Discriminator of the GAN network when the lack of bot samples to train the system makes the deep learning approaches unavailable, showing a superior performance even against bot samples not seen during the training phase. This results suggest that the GAN architecture is a powerful tool not only to generate human-like mouse trajectories, but also to detect bot samples from other synthetic generation methods.}  

As future work, we aim at improving the neuromotor feature set by calculating secondary features inferred from the main ones. Also, we propose \textcolor{black}{to improve the GAN model in two ways: \textit{i)} combine both synthesis methods by using the function-based trajectories as the input of the GAN model instead of Gaussian noise, and \textit{ii)} experimenting with different amount of layers/units in the GAN Generator to increase the variety of the synthetic mouse trajectories generated}. Both techniques could generate more sophisticate and human-like trajectories. Finally, in this paper we only considered mouse trajectories acquired from mouse devices. We also propose to analyze mouse-pad trajectories normally performed when using laptops as another line of research.

\begin{figure}[t!]
\noindent\makebox[\textwidth]{
\centering
\includegraphics[width=0.98\textwidth]{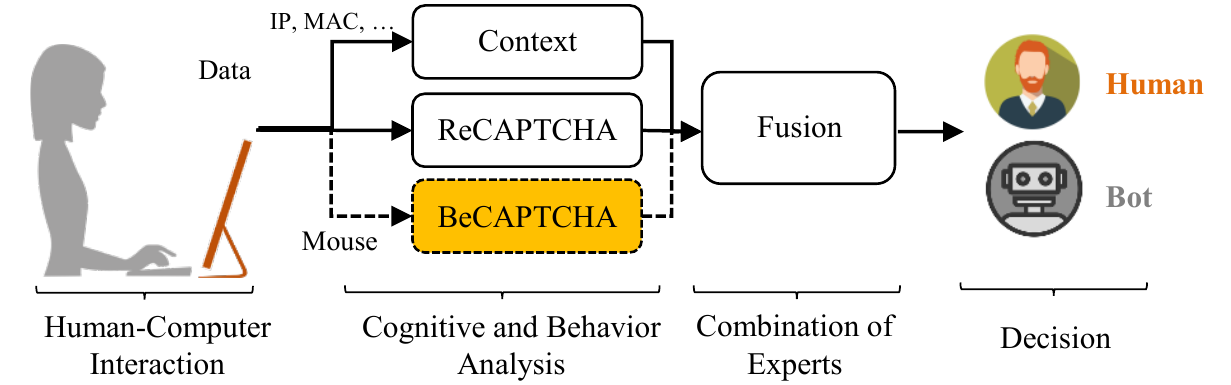}
}
\caption{Block diagram of multimodal bot detection. The response of the bot detector is a combination of responses from different experts. The bot detector proposed in this work can be used independently or in combination with existing bot detectors.}
\label{Fig:BLock_diagram}
\end{figure}

The exploitation of behavioral biometrics for bot detection is an open research line with large opportunities and challenges. These challenges include the study of other ways of interaction beyond mouse such as keystroking \cite{2016_IEEEAccess_KBOC_Aythami,alej2020typenet} or touchscreen gestures \cite{2018_TIFS_Swipe_Fierrez} for bot detection, and their application to mobile scenarios \cite{2021_EAAI_BeCAPTCHA_Acien}. We want to highlight that behavioral CAPTCHAs are compatible with previous CAPTCHA technologies and it could be added as a new cue to improve existing bot detection schemes in a multiple classifier combination \cite{2018_INFFUS_MCSreview1_Fierrez} (see Fig. \ref{Fig:BLock_diagram}). 

Recent fusion techniques incorporating contextual information \cite{2018_INFFUS_MCSreview1_Fierrez} will be also explored for improving BeCAPTCHA. Finally, we'll try to improve our methods taking advantage of existing large-scale human-computer interaction datasets \cite{COMPSAC_ACien} and existing models \cite{2021_AAAI_DeepWriteSYN_Tolosana} by using transfer learning methods \cite{2005_PRL_Fierrez}.

\section*{Acknowledgements}
This work has been supported by projects: TRESPASS-ETN (MSCA-ITN-2019-860813), PRIMA (MSCA-ITN-2019-860315), BIBECA (RTI2018-101248-B-I00 MINECO), and by edBB (UAM). A. Acien is supported by a FPI fellowship from the Spanish MINECO.
\bibliography{paper_v4revisions.bib}
\end{document}